%% file: mckernel.tex
\documentclass[acmtog]{acm}
\usepackage{color}
\usepackage{subfigure}
\usepackage{units}
\usepackage{expl3}
\usepackage{amsmath}
\usepackage{multirow}
\usepackage{colortbl}

\definecolor{Yellow}{rgb}{1,1, 0.6}
\definecolor{Red}{rgb}{1, 0.6, 0.6}
 
\usepackage{booktabs}
\setcitestyle{square}

\renewcommand\footnotetextcopyrightpermission[1]{}
\makeatletter
\renewcommand\@formatdoi[1]{\ignorespaces}
\makeatother

\begin{document}

\title{\huge McKernel: A Library for Approximate Kernel Expansions in Log-linear Time}

\author{J. D. Curt\'o$^{*,1,2,3,4}$, I. C. Zarza$^{*,1,2,3,4}$, F. Yang$^{2,6}$, A. Smola$^{2,5,6}$, F. Torre$^{2,7}$, C. W. Ngo$^{4}$, and L. Gool$^{1}$.}
\affiliation{\institution{\\$^{1}$Eidgen\"ossische Technische Hochschule Z\"urich. $^{2}$Carnegie Mellon.  $^{3}$The Chinese University of Hong Kong. \\ $^{4}$City University of Hong Kong. $^{5}$Amazon. $^{6}$Google. $^{7}$Facebook. \\ $^{*}$Both authors contributed equally.}}

\renewcommand{\shortauthors}{Curt\'o, Zarza, Yang, Smola, Torre, Ngo and Gool.}
\renewcommand{\shorttitle}{McKernel: A Library for Approximate Kernel Expansions in Log-linear Time.}

\authorsaddresses{\{curto,zarza,vangool\}@vision.ee.ethz.ch,fengyang@google.com, \{smola,ftorre\}@cs.cmu.edu,cwngo@cs.cityu.edu.hk \\ \href{https://www.decurto.tw}{decurto.tw} \href{https://www.dezarza.tw}{dezarza.tw}}

\begin{teaserfigure}
\centering
\includegraphics[width=\textwidth]{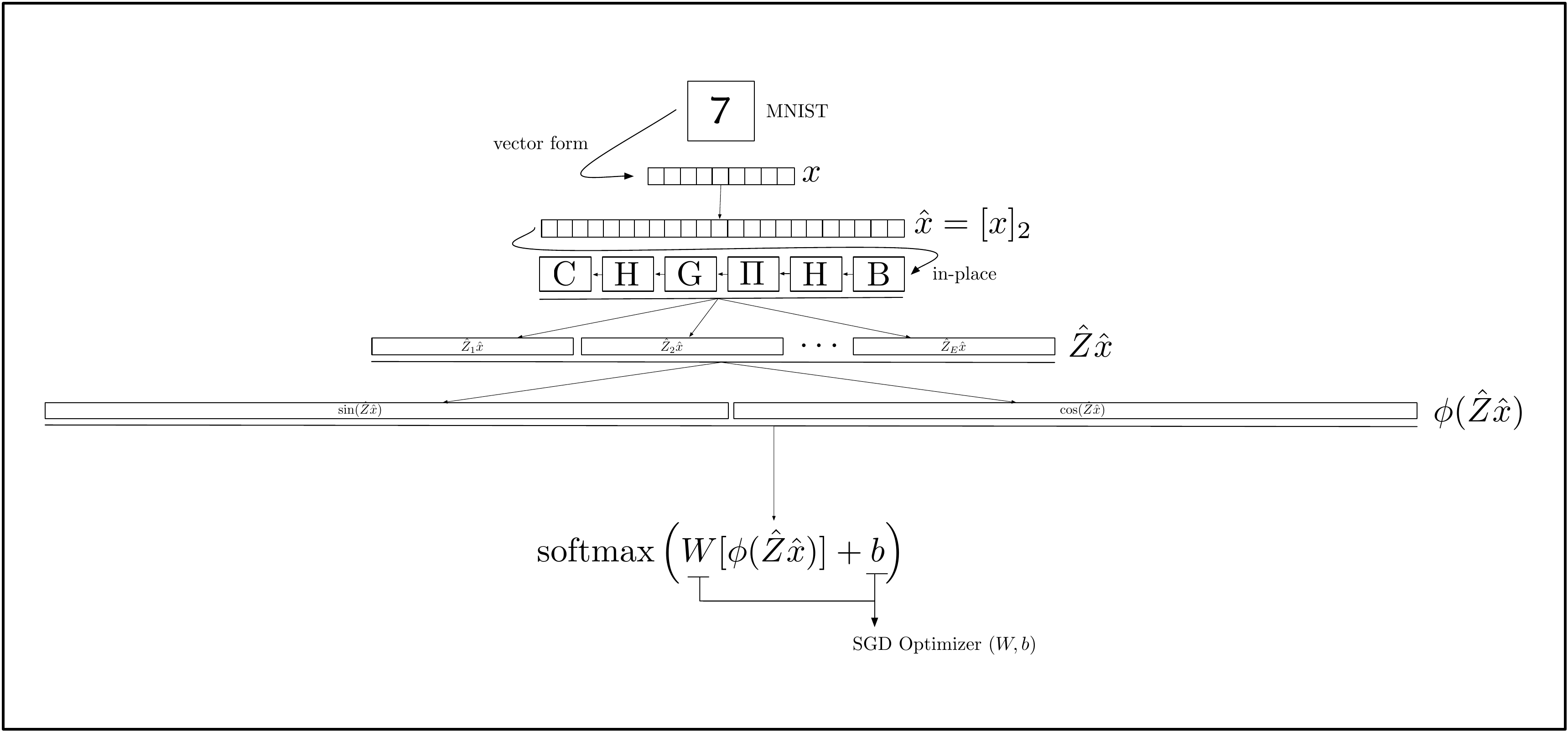}
   \caption{\textbf{Diagram of McKernel}. We visually describe $\textnormal{softmax}(W \tilde{x} + b)$ where $\tilde{x} = \textnormal{mckernel}(x)$. The original image is padded in form of long vector to the nearest power of $2$, mapping $\hat{Z}$ is applied in-place. Calibration $C$ defines the choice of Kernel. The tensor is expanded by the number of Kernel Expansions $E$ building a network with high compositionality. Finally, use real feature map $\phi$, Equation \ref{etn:mcfeatures}. SGD Optimizer finds appropriate weights $W$ and bias $b$. Compute $\hat{Z}$ on-the-fly keeping same seed both for training and testing.}
\label{fgr:t_mckernel}
\end{teaserfigure}

\ExplSyntaxOn
\newcommand\latinabbrev[1]{
  \peek_meaning:NTF . {% Same as \@ifnextchar
    #1\@}%
  { \peek_catcode:NTF a {% Check whether next char has same catcode as \'a, i.e., is a letter
      #1.\@ }%
    {#1.\@}}}
\ExplSyntaxOff

\def\eg{e.g. }
\def\etal{et al. }

%----------------------------------------------------------------------------------------
%	ABSTRACT
%----------------------------------------------------------------------------------------

\begin{abstract}
McKernel introduces a framework to use kernel approximates in the mini-batch setting with Stochastic Gradient Descent (SGD) as an alternative to Deep Learning. Based on Random Kitchen Sinks \cite{Rahimi07}, we provide a C++ library for Large-scale Machine Learning\footnote{McKernel is available at \href{https://www.github.com/curto2/mckernel}{https://www.github.com/curto2/mckernel}}. It contains a CPU optimized implementation of the algorithm in \cite{Le14}, that allows the computation of approximated kernel expansions in log-linear time. The algorithm requires to compute the product of matrices Walsh Hadamard. A cache friendly Fast Walsh Hadamard that achieves compelling speed and outperforms current state-of-the-art methods has been developed. McKernel establishes the foundation of a new architecture of learning that allows to obtain large-scale non-linear classification combining lightning kernel expansions and a linear classifier. It travails in the mini-batch setting working analogously to Neural Networks. We show the validity of our method through extensive experiments on MNIST and FASHION MNIST \cite{Xiao17}.\\ 
\end{abstract}


\maketitle
\fancyfoot{}
\thispagestyle{empty}

%----------------------------------------------------------------------------------------
%	INTRODUCTION
%----------------------------------------------------------------------------------------

\section{Introduction}

Kernel methods offer state-of-the-art estimation performance. They
provide function classes that are flexible and easy to control in
terms of regularization. However, the use of kernels in large-scale
machine learning has been beset with difficulty. This is
because using kernel expansions in large datasets is too expensive in
terms of computation and storage. 
In order to solve this problem, \cite{Le14} proposed an approximation algorithm based on Random Kitchen Sinks by \cite{Rahimi07}, that speeds up the computation of a large range of kernel functions, allowing us to use them in big data. \cite{Rudi17} describes the generalization of Random Features and potential effectiveness. Recent works on the topic build on it to propose state-of-the-art embeddings \cite{Yang15,Moczulski16,Hong17,Kawaguchi18}.
\\

In this work, we go beyond former attempts \cite{Cho09,Wilson16,Al-Shedivat17} and propose a general framework in lieu of Deep Learning. Our goal is to integrate current advances in Neural Networks but at the same time propose a well established theoretically sound background.
\\

\cite{Wigner60} claims the unreasonable effectiveness of mathematics in the natural sciences. \cite{Vapnik18} in the same way, states the unreasonable effectiveness of mathematics in machine learning. Kernel methods originate in rigorous mathematical treatment of the problem, while at the same time are incredibly effective.
\\

At its heart, McKernel requires scalar multiplications, a permutation,
access to trigonometric functions, and two Walsh Hadamard for
implementation. The key computational bottleneck here is the
Walsh Hadamard. We provide a fast, cache friendly SIMD (Single Instruction Multiple Data) oriented implementation
that outperforms state-of-the-art codes such as Spiral \cite{Johnson00}.
To allow for very compact distribution of models, we use hashing
and a Pseudo-random Permutation for portability. 
In this way, for each feature dimension, we only need one floating point number.
\\

In summary, our implementation serves as a drop-in generator of features for
linear methods where attributes are generated on-the-fly \cite{Sharmanska13,Chwialkowski15,Reddi15,Wang15,Li16,Wang17}, such
as for regression, classification, or two-sample tests. This obviates the
need for explicit kernel computations, 
particularly on large amounts of data. 

\paragraph{Outline:}

Learning with Kernels is briefly introduced in Section~\ref{sn:learning}. We begin then with a description of the feature
construction McKernel in Section~\ref{sn:mckernel}. Fast Walsh Hadamard is enunciated in Section~\ref{sn:wh}. This is followed by a
discussion of the computational issues for a SIMD implementation
in Section~\ref{sn:fwh}. The concepts governing the API are
described in Section~\ref{sn:api}. Large-scale Machine Learning by means of McKernel is discussed in Section~\ref{sn:ml}. Concepts regarding TIKHONOV regulation are explained in Section~\ref{sn:tikhonov}. Experimental results can be found in Section~\ref{sn:ea}.
Section~\ref{sn:d} gives a brief overall discussion.

%----------------------------------------------------------------------------------------
%	LEARNING
%----------------------------------------------------------------------------------------

\section{Learning with Kernels}
\label{sn:learning}

The problem of learning \cite{Cortes95,Cucker01,Poggio03,Vapnik09,Vapnik15} arises from the necessity to adapt a model $f: X \to Y$ to a given training set of data $S_{n}=(x_{c},y_{c})_{c=1}^{n}$, with $X \subset \mathbb{R}^{n}$ being closed and $Y \subset \mathbb{R}$, having $f$ good properties of generalization.
\\

Let $(x_{c},y_{c})^{n}_{c=1}$ be the data. Then, we pick a function $k_{x}(x') = k(x,x')$ symmetric, positive definite and continuous on $X \times X$. And set $f: X \to Y$ such that
\begin{align}
f(x) = \sum_{z=0}^{n}t_{z}k_{x_{z}}(x),
\end{align}
where $t=(t_{1},\ldots,t_{n})\in \mathbb{R}^{n}$ and

\begin{align}
(n\gamma I+K)t = y,
\label{etn:learning}
\end{align}

\noindent
where matrix $I$ is the identity, $K$ is the matrix square positive definite with elements $k_{c,r} = k(x_{c},x_{r})$ and $\gamma > 0$ in $\mathbb{R}$.
\\

It turns out this linear system of equations in $n$ variables is well-posed as $K$ is positive and $(n\gamma I+K)$ is strictly positive.
\\

The intuition behind this algorithm, for instance given the Gaussian 
\begin{align}
k(x,x')=\exp \left(-\frac{|| x - x' ||^{2}}{2\sigma^{2}}\right),
\label{etn:rbf}
\end{align}
is that we approximate the unknown function by a weighted superposition of Gaussians, each centered at location $x_{c}$ of one of the $n$ examples. The weight $t_{c}$ of each Gaussian is chosen to minimize the error on the training set. The $\sigma$ of the Gaussian, together with $\gamma$, controls the degree of smoothing, of noise tolerance and generalization.
\\

\cite{Vapnik18} proposes a generalization of this framework by the use of invariants. Equation \ref{etn:learning} becomes 
\begin{align}
(n\gamma I+VK)t = Vy,
\label{etn:learning2}
\end{align}
where $V$ takes into account mutual positions of observed vectors and elements $V(c,z)$ can be computed for high-dimensional problems as follows
\begin{align}
V(c,z) = \sum^{d}_{k=1} \left( t_{k} - \max (x^{k}_{c}, x^{k}_{z}) \right),
\label{etn:learning3}
\end{align}
with $0 \leq x_{k} \leq t_{k}$. Matrix $V$ is a symmetric non-negative matrix. The main idea is to take into account that the desired decision rule is related to the conditional probability function of the observations.

%----------------------------------------------------------------------------------------
%	MCKERNEL
%----------------------------------------------------------------------------------------

\section{McKernel}
\label{sn:mckernel}

Kernel methods work by defining a kernel function $k(x,x')$ on a
domain $X$. We can write $k$ as inner product between feature maps, as follows 

\begin{align}
    k(x,x') = \langle \phi(x),\phi(x')\rangle
\end{align}
for some suitably chosen $\phi$. Random Kitchen Sinks \cite{Rahimi07}
approximate this mapping of features $\phi$ by a FOURIER expansion in the
case of radial basis function (RBF), scilicet whenever
$k(x,x') = \kappa(x-x')$. This is possible since the FOURIER transform
diagonalizes the corresponding integral operator. This leads to 
\begin{align}
    k(x,x') = \int \exp(i \langle w,x\rangle) \exp(-i \langle w,x'\rangle) d\rho(w)
\end{align}
for some $L_2$ measurable function $\rho(\omega) \geq 0$ that is given by
the FOURIER transform of $\kappa$. Random Kitchen Sinks exploit this
by replacing the integral by sampling $\omega \sim \rho(\omega)/\|\rho\|_1$. This allows for
finite dimensional expansions but it is costly due to the large number
of inner products required. \cite{Le14} resolves this for rotationally
invariant $\kappa$ by providing a fast approximation of the matrix $W
:= [\omega_1, \ldots \omega_n]$. 

This is best seen for the RBF kernel, Equation \ref{etn:rbf}. Since FOURIER
transforms of Gaussians are Gaussians, albeit with inverse covariance,
it follows that $\rho(\omega) \propto \exp \left(-\frac{\sigma^2}{2}
  \|\omega\|^2\right)$ and that $W$ contains random
variables independent and identically distributed (i.i.d.) Gaussian. It is this matrix that McKernel approximates via 
\begin{align}
  \hat{Z} := \frac{1}{\sigma \sqrt{n}} C H G \Pi H B.
  \label{etn:mckernel}
\end{align}
Here $C, G$ and $B$ are diagonal matrices, $\Pi$ is a random
permutation matrix and $H$ is the Hadamard. Whenever the number
of rows in $W$ exceeds the dimensionality of the data, we can simply
generate multiple instances of $\hat{Z}$, drawn i.i.d., until the required
number of dimensions is obtained.\\

\begin{description}
\item[Binary $B$.] This is a matrix with entries
  $B_{kk} \in \{\pm 1\}$, drawn from the uniform distribution. To
  avoid memory footprint, we simply use
  Murmurhash as
  function of hashing and extract bits from $h(k,x)$ with
  $x \in \{0, \ldots N\}$.
  \\
\item[Hadamard $H$.] This matrix is iteratively composed
  of $H_{n} =\begin{bmatrix}
H_{n-1} & H_{n-1}\\
H_{n-1} & -H_{n-1}
\end{bmatrix}$. It is fixed
  and matrix-vector products are carried out efficiently in
  $O(n \log n)$ time using the Fast Walsh Hadamard. We will discuss implementation details for a fast variant in Section \ref{sn:fwh}.
  \\
\item[Permutation $\Pi$.] We generate a random permutation using
  the FISHER-YATES shuffle. That is, given a list $L = \{1, \ldots n\}$ we
  generate permutations recursively as follows: pick a random element
  from $L$. Use this as the image of $n$ and move $n$ to the position
  where the element was removed. The algorithm runs in linear time and
  its coefficients can be stored in $O(n)$ space. Moreover, to obtain
  a deterministic mapping, replace the generator of random numbers with
  calls to the function of hashing. 
  \\
\item[Gaussian $G$.] This is a matrix diagonal with entries i.i.d.
  Gaussian. We generate the random variates using the
  BOX-MULLER transform \cite{Box58} while substituting the generator of random numbers
  by calls to the function of hashing to allow us to
  recompute the values at any time without the need to store random
  numbers.
   \\
\item[Calibration $C$.] This is a random scaling operator whose
  behavior depends on the type of kernel chosen, such as the RBF MAT\'ERN
  Kernel, the RBF Kernel or any other radial spectral
  distribution \cite{Yang14}.\\
\end{description}

Ultimately, compute the feature pairs by assigning
\begin{align}
  [\cos(\hat{Z}x),\sin(\hat{Z}x)]. 
    \label{etn:mcfeatures}
\end{align}

In particular, McKernel computes the features by using the real version of the complex feature map $\phi$ in \cite{Rahimi07}. SIMD vectorized instructions and cache locality are used to increase speed performance. These allow a speed improvement of 18x times for a $2^{24}$ dimension input matrix.

%----------------------------------------------------------------------------------------
%	FAST WALSH HADAMARD
%----------------------------------------------------------------------------------------

\section{Fast Walsh Hadamard}
\label{sn:wh}

Applications of the WALSH HADAMARD transform range across several areas, including Machine Learning \cite{Lu13,Andoni15} and Computer Vision \cite{Ben-Artzi07,Ouyang10}. 
\\

A na\"ive implementation results in complexity $O(n^{2})$. A divide-and-conquer approach for this task that runs in time $O(n \log n)$ can be derived as follows.
\\

We define the matrix Hadamard $H_{n}$
\begin{align}
H_{0}=[1] \\
H_{n}=
\begin{bmatrix}
H_{n-1} & H_{n-1}\\
H_{n-1} & - H_{n-1}
\end{bmatrix},
\end{align}
with dimension $2^{n} \times 2^{n}$.
\\

We want to compute the product of matrix vector $H_{n}\cdot c$, being $c =(c_{o}, c_{1})$ where $|c_{0}|=|c_{1}| = \frac{|c|}{2}$,
\begin{align}
H_{n}\cdot c=
\begin{bmatrix}
H_{n-1} c_{0} + H_{n-1}c_{1}\\
H_{n-1} c_{0}  - H_{n-1}c_{1}
\end{bmatrix}.
\end{align}
Hence, we only need to compute recursively $H_{n-1}c_{0}$ and $H_{n-1}c_{1}$ to obtain $H_{n}\cdot c$ via additions and subtractions. The running time is 
\begin{align}
T(n) = 2T(n/2)+O(n),
\end{align}
that gives $T(n)=O(n\log n)$.
\\

In McKernel we generalize this approach to compute the resulting matrix Hadamard from a hard-coded specific-size routine.

\section{Implementation of Fast Walsh Hadamard}
\label{sn:fwh}

A key part of the library is an efficient
implementation of Fast Walsh Hadamard. In particular,
McKernel offers considerable improvement over
Spiral, due to automatic
code generation, the use of SIMD intrinsics (SSE2 using 128 bit
registers) and loop unrolling. This decreases the memory overhead. \\

McKernel proceeds with vectorized sums and subtractions iteratively for the first $\frac{n}{2^z}$ input vector positions (where $n$ is the length of the input vector and $z$ the iteration starting from $1$), computing the intermediate operations of the COOLEY-TUKEY algorithm till a small routine Hadamard that fits in cache. Then the algorithm continues in the same way but starting from the smallest length and doubling on each iteration the input dimension until the whole Fast Walsh Hadamard is done in-place. 
\\

\begin{figure}[H]
\centering
\includegraphics[scale=1]{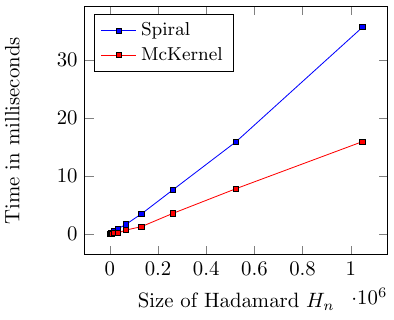}
\caption{\textbf{Comparison of Fast Walsh Hadamard}. McKernel (red) outperforms Spiral (blue) across the range of arguments.}
\label{fgr:fwh}
\end{figure}

For instance, on an intel i5-4200 CPU @ 1.60GHz laptop the performance obtained can be observed in Figure \ref{fgr:fwh}.

Our code outperforms Spiral consistently throughout
the range of arguments, see Table \ref{tbl:mckernel}. Furthermore, Spiral needs to precompute trees
and by default can only perform the computation up to matrix size $n =
2^{20}$. On the other hand, our implementation works for any given size
since it computes the high-level partitioning dynamically. 

\begin{table}[t]
\caption{Numeric Comparison of Fast Walsh Hadamard.}
\label{tbl:mckernel}
\vskip 0.15in
\begin{center}
\begin{small}
\begin{sc}
\begin{tabular}{lll}
\toprule
   $|H_{n}|$ & McKernel $t(ms)$ & Spiral $t(ms)$\\
   
       \midrule
        1024 & 0 & 0.0333\\
        2048 & 0.0333 & 0.0667\\
        4096 & 0.1 & 0.167\\
        8192 & 0.0667 & 0.2\\
        16384 & 0.2 & 0.467\\
        32768 & 0.2 & 0.9\\
        65536 & 0.7 & 1.667\\
        131072 & 1.3 & 3.5\\
        262144 & 3.6 & 7.667\\
        524288 & 7.86 & 15.9667\\
        1048576 & 15.9667 & 35.7\\
\bottomrule
\end{tabular}
\end{sc}
\end{small}
\end{center}
\vskip -0.1in
\end{table}

%----------------------------------------------------------------------------------------
%	API
%----------------------------------------------------------------------------------------

\section{API Description}
\label{sn:api}

The API follows the design
pattern in factory. That is, while the object McKernel is fairly generic in terms
of computation, we have a factory that acts as a means of
instantiating the parameters according to pre-specified sets of
parameters, e.g. a RBF Kernel or a RBF MAT\'ERN Kernel. The so-chosen
parameters are deterministic, given by the values of a function of hashing. The advantage of this approach is that there is no need to
save the coefficients generated for McKernel when deploying the
functions.
\\

McKernel is integrated into a fully-fledged C++ DL framework that lets the user experiment, among other things, with dropout, convolutions, different activation functions, layer normalization, maxpooling, L1 and L2 regularization, gradient clipping, autoencoders, residual blocks, SGD optimization with momentum and dataset loading. That said, it also includes some classical algorithms for learning such as linear and logistic regression.

%----------------------------------------------------------------------------------------
%	CUSTOMIZING MCKERNEL
%----------------------------------------------------------------------------------------

\subsection{Customizing McKernel}

For instance, to generate each $C_{kk}$ entry for RBF MAT\'ERN Kernel we draw $t$ i.i.d. samples from the $n$-dimensional unit ball $S_{n}$, add them and compute its Euclidean norm. 
\\

To draw efficiently samples from $S_{n}$ we use the algorithm provided below.
\\

Let $ X = (X_{1}, \ldots, X_{n})$ be a vector of i.i.d. random variables drawn from $N(0,1)$, and let $||X||$ be the Euclidean norm of $X$, then $Y = \left(\frac{X_{1}}{||X||}, \ldots, \frac{X_{n}}{||X||}\right)$ is uniformly distributed over the $n$-sphere, where $Y$ is the projection of $X$ onto the surface of the $n$-dimensional sphere. To draw uniform random variables in the $n$-ball, we multiply $Y$ by $U^{1/n}$ where $U \sim  U(0,1)$. This can be proved as follows: Let $Z=(Z_{1}, \ldots, Z_{n})$ be a random vector uniformly distributed in the unit ball. Then, the radius $R = || Z ||$ satisfies $P(R \leq r) = r^{n}$. Now, by the inverse transform method we get $R= U^{1/n}$. Therefore to sample uniformly from the $n$-ball the following algorithm is used: 
\begin{align}
Z = r U^{1/n} \frac{X}{||X||} .
\end{align}

%----------------------------------------------------------------------------------------
%	ML
%----------------------------------------------------------------------------------------

\section{Large-scale Machine Learning}
\label{sn:ml}
We introduce McKernel as an alternative to Deep Learning, where we argue that current techniques of Neural Networks are surrogates to very large kernel expansions, where optimization is done in a huge parameter space where the majority of learned weights are trivial to the actual problem statement.
\\

We present here the idea that current developments in very deep neural networks can be achieved while drastically reducing the number of parameters learned. We build on the work in \cite{Rahimi07} and \cite{Le14} to expand its scope to mini-batch training with SGD Optimizer. Our concern is to demonstrate that we are able to get the same gains obtained in Deep Learning by the use of McKernel and a linear classifier but with a behemoth kernel expansion. 
\\

Current research in Neural Networks is over-optimizing parameters that are indeed not informative for the problem to solve. That is, we pioneer the notion that Deep Learning is learning the inner parameters of a very large kernel expansion and, in the end, is doing a brute-force search of the appropriate kernel $k$ thats fits well the data.
\\

Observe that McKernel generates pseudo-random numbers by the use of hashing which is key for large-scale data. It allows to obtain a deterministic behavior and at the same time to load the weights on both training and testing without the need to actually store the matrices. As a further matter, it is crucial for distributed computation.
\\

Notice here that the fact that we can increase the number of kernel expansions building highly hierarchical networks, see Equations \ref{etn:mckernel} and \ref{etn:mcfeatures}, gives the property of compositionality to McKernel. Namely, the theoretical guarantee to avoid the curse of dimensionality \cite{Poggio17, Mhaskar17}.
\\

We can behold that McKernel is corresponding to networks of the form
\begin{align}
G(x) = \sum^{N}_{k=1} a_{k} \exp \left( - |x - x_{k}|^{2} \right), \qquad x \in \mathbb{R}^{d}.
\end{align}

In the following section we build on these ideas to propose some very simple examples to illustrate the essence of the problem and the solution proposed.

%----------------------------------------------------------------------------------------
%	TIKHONOV Regularization
%----------------------------------------------------------------------------------------

\section{TIKHONOV Regularization}
\label{sn:tikhonov}

Let $H$ be the hypothesis space of functions, in the problem of Empirical Risk Minimization (ERM) we want to find $f \in H$ that minimizes

\begin{align}
\frac{1}{n}\sum^{n}_{c=1}(f(x_{c}) - y_{c})^{2}.
\end{align}

This problem is in general ill-posed, depending on the choice of $H$. Following Tikhonov \cite{Girosi95,Girosi98} we minimize instead over the hypothesis space $H_{K}$, the regularized functional

\begin{align}
\frac{1}{n}\sum^{n}_{c=1}(f(x_{c}) - y_{c})^{2} + \lambda ||f||^{2}_{K},
\end{align}

\noindent
where $||f||^{2}_{K}$ is the norm in $H_{K}$ - the REPRODUCING KERNEL HILBERT Space defined by the kernel K.
\\

In general, under the TIKHONOV regularization scheme that follows,
\begin{align}
\min_{w \in \mathbb{R}^{D}} \hat{E}(f_{w}) + \lambda ||w||^{2},
\label{etn:tikhonov}
\end{align}

\noindent
where $||w||^{2}$ is the regularizer and controls the stability of the solution and $\lambda$ balances the error term and the regularizer. Different classes of methods are determined by the appropriate choice of loss function in Equation \ref{etn:tikhonov}. Here we consider
\begin{align}
\hat{E}(f_{w}) = \frac{1}{n} \sum^{n}_{c=1} l(y_{c},f_{w}(x_{c}))
\end{align}
\noindent
with loss function $l$ defined as
\begin{align}
l(y,f_{w}(x)) = \log (1 + \exp(-y f_{w}(x))),
\label{etn:loss}
\end{align}
\noindent
videlicet Logistic Regression.
\\

Considering the logistic loss is differentiable and that we are in a large-scale setting a reasonable candidate to compute a minimizer is the Stochastic Gradient Descent (SGD),
\begin{align}
w_{t+1} = w_{t} - \gamma \Delta g_{c_{t}}(w_{t}),
\end{align}
where $c_{t}$ denotes a stochastic sequence of indices and $\gamma$ is the learning rate.
\\

In this line of argument, \cite{Liang18,Kawaguchi19,Sohl-Dickstein19} state that local minimization is well posed in Deep Learning using SGD.
\\

Augmenting the number of kernel expansions, and thus the representational power of the model, gives a degree of over-parametrization. That is to say, we increase the size of the network to fit the training data. Given these constraints, B\'EZOUT theorem argues the existence of a large number of degenerate global minimizers with zero empirical error, that are very likely to be found by SGD that will in addition select with higher probability the most robust zero-minimizer \cite{Liao17}. 

%----------------------------------------------------------------------------------------
%	EMPIRICAL ANALYSIS AND EXPERIMENTS
%----------------------------------------------------------------------------------------

\section{Empirical Analysis and Experiments} 
\label{sn:ea}

We generalize the use of McKernel in mini-batch with SGD Optimizer, Figure \ref{fgr:t_mckernel}, drastically reducing the number of parameters that need to be learned to achieve comparable state-of-the-art results to Deep Learning. 
\\

What is more, current breakthroughs in Neural Networks can be easily derived from Equation \ref{etn:mckernel}. Say for instance, Batch Normalization \cite{Ioffe15} can be obtained from the normalizing factor. Or for example, the use of ensembles \cite{Lakshminarayanan17} and multi-branch architectures \cite{Zhang18} to improve performance can be seen on Figure \ref{fgr:t_mckernel}, as it follows from the increase on Kernel Expansions. Not only that, but increasing the number of Kernel Expansions has another great property; data augmentation \cite{Tran17,Cubuk18}, which has recently seen a lot of interest. Its importance follows directly from the construction of McKernel, take the data, apply slightly (randomized) different functions to it to create new high-dimensional samples that will aid the process of learning. It also explains the need for backpropagation \cite{Lecun98}: certain types of data will work better for certain kernels, so it may be necessary to learn the appropriate Calibration $C$ and $G$ that fit well the data. Besides, learning $B$ acts as mechanism of attention \cite{Bahdanau15,Luong15,Vaswani17}. Further, dropout \cite{Srivastava14} follows directly from the use of the Subsampled Randomized Hadamard. Additionally, research on finding alternate activation functions \cite{Maas13,He15,Clevert16,Klarbauer17} can be deduced from looking for different mappings in Equation \ref{etn:mcfeatures}. 
\\

Note here that the number of parameters to be estimated is of the order of thousands, proportional to the number of classes (depending on the size of the input image and the number of kernel expansions),

\begin{align}
C \cdot (2 \cdot [S]_{2} \cdot E + 1),
\end{align}

\noindent
where $C$ is the number of classes, $[\cdot]_{2}$ is an operator that returns the next power of $2$, $S$ is the size of the input samples and $E$ is the number of Kernel Expansions. A drastic reduction compared to Neural Networks, while achieving comparable performance. Training time is therefore severely reduced and SVM kernel like learning can be achieved at scale.

\begin{figure}[H]
\centering
\includegraphics[scale=0.66]{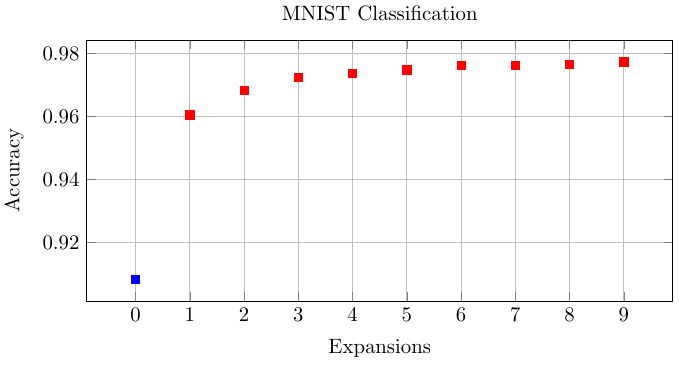}
\caption{\textbf{MNIST Classification.} Logistic Regression (blue) and RBF MAT\'ERN (red) with increasing number of Kernel Expansions. $32768$ samples of training data and $8192$ samples of testing data are used in learning. RBF MAT\'ERN hyper-parameters, $\sigma=1.0$, $t=40$. Seed $1398239763$, learning rate $\gamma = 0.001$ and batch size $10$. LR learning rate $0.01$. Number of epochs $20$.
}
\label{fgr:rbfmatern}
\end{figure}

SGD Optimizer finds $W$ and $b$ in
\begin{align}
\textnormal{softmax} \left(W[\phi(\hat{Z}\hat{x})] + b\right),
\end{align}
where $\phi=\left(\sin(\cdot),\cos(\cdot)\right)$, $\hat{x}=[x]_{2}$. Namely, it minimizes the loss $l$ in Equation \ref{etn:loss}.

\begin{figure}[H]
\centering
\includegraphics[scale=0.66]{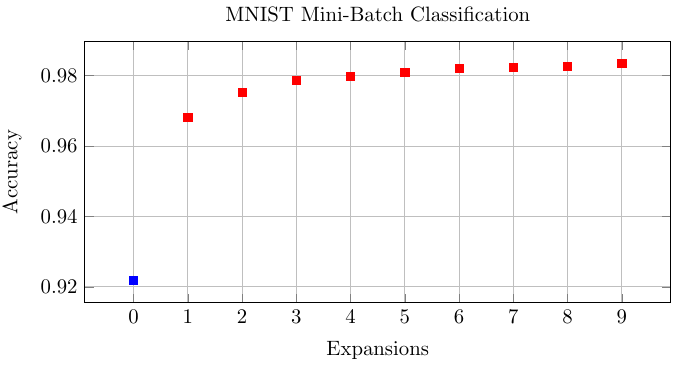}
\caption{\textbf{MNIST Mini-Batch Classification.} Logistic Regression (LR) (blue) and RBF MAT\'ERN (red) with increasing number of Kernel Expansions. $60000$ samples of training data and $10000$ samples of testing data are used in learning. RBF MAT\'ERN hyper-parameters, $\sigma=1.0$, $t=40$. Seed $1398239763$, learning rate $\gamma = 0.001$ and batch size $10$. LR learning rate $0.01$. Number of epochs $20$.
}
\label{fgr:rbfmatern2}
\end{figure}

Figures \ref{fgr:rbfmatern} and \ref{fgr:rbfmatern2} show RBF MAT\'ERN, $\textnormal{softmax}(W \tilde{x} + b)$ where $\tilde{x} = \textnormal{mckernel}(x)$, performance compared to logistic regression, $\textnormal{softmax}(Wx + b)$, in full-batch and mini-batch on MNIST, respectively. A fixed seed is used to obtain deterministic reproducible behavior. In full-batch, the number of samples for train and test is rounded to the nearest power of $2$ due to algorithm constraint.
\\

The same kind of intuition that applies to Neural Networks, where the deeper the network, the better the results, holds. But this time depending on the number of kernel expansions used. 
\\

FASHION MNIST \cite{Xiao17} is similar in scope to MNIST but relatively more difficult. Instead of classifying digits, we focus now on the task of fashion. It consists on ten classes of clothing; T-shirt/top, Trouser, Pullover, Dress, Coat, Sandal, Shirt, Sneaker, Bag and Ankle boot. 
\\

\begin{figure}[H]
\centering
\includegraphics[scale=0.66]{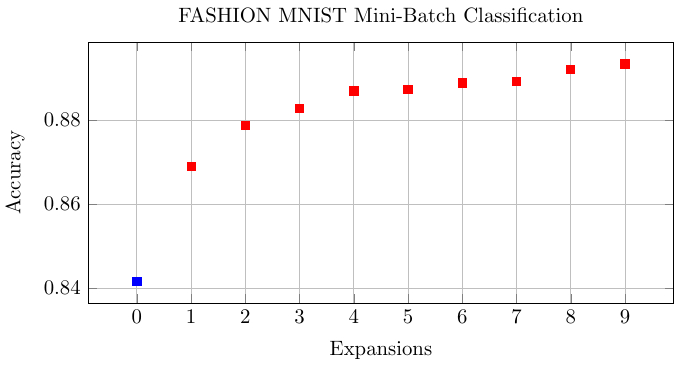}
\caption{\textbf{FASHION MNIST Mini-Batch Classification.} Logistic Regression (LR) (blue) and RBF MAT\'ERN (red) with increasing number of Kernel Expansions. $60000$ samples of training data and $10000$ samples of testing data are used in learning. RBF MAT\'ERN hyper-parameters, $\sigma=1.0$, $t=40$. Seed $1398239763$, learning rate $\gamma = 0.001$ and batch size $10$. LR learning rate $0.01$. Number of epochs $20$.
}
\label{fgr:fashion_mnist}
\end{figure}

Figure \ref{fgr:fashion_mnist} shows RBF MAT\'ERN performance compared to logistic regression in mini-batch. Comparable state-of-the-art performance to Deep Learning is achieved. The model presents a similar behavior to the one seen in MNIST dataset. McKernel performs analogously to modern techniques in Neural Networks in this highly non-linear problem of estimation. 

%----------------------------------------------------------------------------------------
%	DISCUSSION
%----------------------------------------------------------------------------------------

\section{Discussion} 
\label{sn:d}

In this manuscript we provide a new framework of learning and illustrate with two examples that achieves comparable state-of-the-art performance to Neural Networks, proposing a new way to understand Deep Learning, as a huge kernel expansion where optimization is only performed over the parameters that are actually relevant to the problem at-hand. At the same time, a new methodology to build highly compositional networks for Large-scale Machine Learning is introduced.
\\

We account for both the theoretical underpinnings and the practical implications to establish the building blocks of a unifying theory of learning.

%----------------------------------------------------------------------------------------
%	BIBLIOGRAPHY
%----------------------------------------------------------------------------------------

\bibliographystyle{ACM-Reference-Format}
\input{mckernel.bbl}

\end{document}

%% file: mckernel.bbl

%% file: mckernel.bbl
\begin{thebibliography}{53}

%%% the Web address.

\ifx \showCODEN    \undefined \def \showCODEN     #1{\unskip}     \fi
\ifx \showDOI      \undefined \def \showDOI       #1{#1}\fi
\ifx \showISBNx    \undefined \def \showISBNx     #1{\unskip}     \fi
\ifx \showISBNxiii \undefined \def \showISBNxiii  #1{\unskip}     \fi
\ifx \showISSN     \undefined \def \showISSN      #1{\unskip}     \fi
\ifx \showLCCN     \undefined \def \showLCCN      #1{\unskip}     \fi
\ifx \shownote     \undefined \def \shownote      #1{#1}          \fi
\ifx \showarticletitle \undefined \def \showarticletitle #1{#1}   \fi
\ifx \showURL      \undefined \def \showURL       {\relax}        \fi
% invisible to TeX
\providecommand\bibfield[2]{#2}
\providecommand\bibinfo[2]{#2}
\providecommand\natexlab[1]{#1}
\providecommand\showeprint[2][]{arXiv:#2}

\bibitem[\protect\citeauthoryear{Al-Shedivat, Wilson, Saatchi, Hu, and
  Xing}{Al-Shedivat et~al\mbox{.}}{2017}]%
        {Al-Shedivat17}
\bibfield{author}{\bibinfo{person}{M. Al-Shedivat}, \bibinfo{person}{A.~G.
  Wilson}, \bibinfo{person}{Y. Saatchi}, \bibinfo{person}{Z. Hu}, {and}
  \bibinfo{person}{E.~P. Xing}.} \bibinfo{year}{2017}\natexlab{}.
\newblock \showarticletitle{Learning scalable deep kernels with recurrent
  structure}.
\newblock \bibinfo{journal}{\emph{JMLR}} (\bibinfo{year}{2017}).
\newblock


\bibitem[\protect\citeauthoryear{Andoni, Indyk, Laarhoven, Razenshteyn, and
  Schmidt}{Andoni et~al\mbox{.}}{2015}]%
        {Andoni15}
\bibfield{author}{\bibinfo{person}{A. Andoni}, \bibinfo{person}{P. Indyk},
  \bibinfo{person}{T. Laarhoven}, \bibinfo{person}{I. Razenshteyn}, {and}
  \bibinfo{person}{L. Schmidt}.} \bibinfo{year}{2015}\natexlab{}.
\newblock \showarticletitle{Practical and Optimal LSH for Angular Distance}.
\newblock \bibinfo{journal}{\emph{NIPS}} (\bibinfo{year}{2015}).
\newblock


\bibitem[\protect\citeauthoryear{Bahdanau, Cho, and Bengio}{Bahdanau
  et~al\mbox{.}}{2015}]%
        {Bahdanau15}
\bibfield{author}{\bibinfo{person}{D. Bahdanau}, \bibinfo{person}{K. Cho},
  {and} \bibinfo{person}{Y. Bengio}.} \bibinfo{year}{2015}\natexlab{}.
\newblock \showarticletitle{Neural Machine Translation by Jointly Learning to
  Align and Translate}.
\newblock \bibinfo{journal}{\emph{ICLR}} (\bibinfo{year}{2015}).
\newblock


\bibitem[\protect\citeauthoryear{Ben-Artzi, Hel-Or, and Hel-Or}{Ben-Artzi
  et~al\mbox{.}}{2007}]%
        {Ben-Artzi07}
\bibfield{author}{\bibinfo{person}{G. Ben-Artzi}, \bibinfo{person}{H. Hel-Or},
  {and} \bibinfo{person}{Y. Hel-Or}.} \bibinfo{year}{2007}\natexlab{}.
\newblock \showarticletitle{The Gray-Code Filter Kernels}.
\newblock \bibinfo{journal}{\emph{TPAMI}} (\bibinfo{year}{2007}).
\newblock


\bibitem[\protect\citeauthoryear{Box and Muller}{Box and Muller}{1958}]%
        {Box58}
\bibfield{author}{\bibinfo{person}{G.~E.~P. Box} {and} \bibinfo{person}{M.~E.
  Muller}.} \bibinfo{year}{1958}\natexlab{}.
\newblock \showarticletitle{A Note on the Generation of Random Normal
  Deviates}.
\newblock \bibinfo{journal}{\emph{The Annals of Mathematical Statistics}}
  (\bibinfo{year}{1958}).
\newblock


\bibitem[\protect\citeauthoryear{Cho and Saul}{Cho and Saul}{2009}]%
        {Cho09}
\bibfield{author}{\bibinfo{person}{Y. Cho} {and} \bibinfo{person}{L.~K. Saul}.}
  \bibinfo{year}{2009}\natexlab{}.
\newblock \showarticletitle{Kernel Methods for Deep Learning}.
\newblock \bibinfo{journal}{\emph{NIPS}} (\bibinfo{year}{2009}).
\newblock


\bibitem[\protect\citeauthoryear{Chwialkowski, Ramdas, Sejdinovic, and
  Gretton}{Chwialkowski et~al\mbox{.}}{2015}]%
        {Chwialkowski15}
\bibfield{author}{\bibinfo{person}{K. Chwialkowski}, \bibinfo{person}{A.
  Ramdas}, \bibinfo{person}{D. Sejdinovic}, {and} \bibinfo{person}{A.
  Gretton}.} \bibinfo{year}{2015}\natexlab{}.
\newblock \showarticletitle{Fast Two-Sample Testing with Analytic
  Representations of Probability Measures}.
\newblock \bibinfo{journal}{\emph{NIPS}} (\bibinfo{year}{2015}).
\newblock


\bibitem[\protect\citeauthoryear{Clevert, Unterthiner, and Hochreiter}{Clevert
  et~al\mbox{.}}{2016}]%
        {Clevert16}
\bibfield{author}{\bibinfo{person}{D. Clevert}, \bibinfo{person}{T.
  Unterthiner}, {and} \bibinfo{person}{S. Hochreiter}.}
  \bibinfo{year}{2016}\natexlab{}.
\newblock \showarticletitle{Fast and Accurate Deep Network Learning by
  Exponential Linear Units ({ELU}s)}.
\newblock \bibinfo{journal}{\emph{ICLR}} (\bibinfo{year}{2016}).
\newblock


\bibitem[\protect\citeauthoryear{Cortes and Vapnik}{Cortes and Vapnik}{1995}]%
        {Cortes95}
\bibfield{author}{\bibinfo{person}{C. Cortes} {and} \bibinfo{person}{V.
  Vapnik}.} \bibinfo{year}{1995}\natexlab{}.
\newblock \showarticletitle{Support vector networks}.
\newblock \bibinfo{journal}{\emph{Machine Learning}} (\bibinfo{year}{1995}).
\newblock


\bibitem[\protect\citeauthoryear{Cubuk, Zoph, Mane, Vasudevan, and Le}{Cubuk
  et~al\mbox{.}}{2018}]%
        {Cubuk18}
\bibfield{author}{\bibinfo{person}{E.~D. Cubuk}, \bibinfo{person}{B. Zoph},
  \bibinfo{person}{D. Mane}, \bibinfo{person}{V. Vasudevan}, {and}
  \bibinfo{person}{Q.~V. Le}.} \bibinfo{year}{2018}\natexlab{}.
\newblock \showarticletitle{AutoAugment: Learning Augmentation Policies from
  Data}.
\newblock \bibinfo{journal}{\emph{arXiv:1805.09501}} (\bibinfo{year}{2018}).
\newblock


\bibitem[\protect\citeauthoryear{Cucker and Smale}{Cucker and Smale}{2001}]%
        {Cucker01}
\bibfield{author}{\bibinfo{person}{F. Cucker} {and} \bibinfo{person}{S.
  Smale}.} \bibinfo{year}{2001}\natexlab{}.
\newblock \showarticletitle{On the mathematical foundations of learning}.
\newblock \bibinfo{journal}{\emph{Bulletin of the American Mathematical Society
  (AMS)}} (\bibinfo{year}{2001}).
\newblock


\bibitem[\protect\citeauthoryear{Girosi}{Girosi}{1998}]%
        {Girosi98}
\bibfield{author}{\bibinfo{person}{F. Girosi}.}
  \bibinfo{year}{1998}\natexlab{}.
\newblock \showarticletitle{An equivalence between sparse approximation and
  Support Vector Machines}.
\newblock \bibinfo{journal}{\emph{Neural Computation}} (\bibinfo{year}{1998}).
\newblock


\bibitem[\protect\citeauthoryear{Girosi, Jones, and Poggio}{Girosi
  et~al\mbox{.}}{1995}]%
        {Girosi95}
\bibfield{author}{\bibinfo{person}{F. Girosi}, \bibinfo{person}{M. Jones},
  {and} \bibinfo{person}{T. Poggio}.} \bibinfo{year}{1995}\natexlab{}.
\newblock \showarticletitle{Regularization theory and neural networks
  architectures}.
\newblock \bibinfo{journal}{\emph{Neural Computation}} (\bibinfo{year}{1995}).
\newblock


\bibitem[\protect\citeauthoryear{He, Zhang, Ren, and Sun}{He
  et~al\mbox{.}}{2015}]%
        {He15}
\bibfield{author}{\bibinfo{person}{K. He}, \bibinfo{person}{X. Zhang},
  \bibinfo{person}{S. Ren}, {and} \bibinfo{person}{J. Sun}.}
  \bibinfo{year}{2015}\natexlab{}.
\newblock \showarticletitle{Delving Deep into Rectifiers: Surpassing
  Human-Level Performance on ImageNet Classification}.
\newblock \bibinfo{journal}{\emph{ICCV}} (\bibinfo{year}{2015}).
\newblock


\bibitem[\protect\citeauthoryear{Hong, Yuan, and Bhattacharjee}{Hong
  et~al\mbox{.}}{2017}]%
        {Hong17}
\bibfield{author}{\bibinfo{person}{W. Hong}, \bibinfo{person}{J. Yuan}, {and}
  \bibinfo{person}{S.~D. Bhattacharjee}.} \bibinfo{year}{2017}\natexlab{}.
\newblock \showarticletitle{Fried Binary Embedding for High-Dimensional Visual
  Features}.
\newblock \bibinfo{journal}{\emph{CVPR}} (\bibinfo{year}{2017}).
\newblock


\bibitem[\protect\citeauthoryear{Ioffe and Szegedy}{Ioffe and Szegedy}{2015}]%
        {Ioffe15}
\bibfield{author}{\bibinfo{person}{S. Ioffe} {and} \bibinfo{person}{C.
  Szegedy}.} \bibinfo{year}{2015}\natexlab{}.
\newblock \showarticletitle{Batch normalization: Accelerating deep network
  training by reducing internal covariate shift}.
\newblock \bibinfo{journal}{\emph{ICML}} (\bibinfo{year}{2015}).
\newblock


\bibitem[\protect\citeauthoryear{Johnson and P{\"u}schel}{Johnson and
  P{\"u}schel}{2000}]%
        {Johnson00}
\bibfield{author}{\bibinfo{person}{J. Johnson} {and} \bibinfo{person}{M.
  P{\"u}schel}.} \bibinfo{year}{2000}\natexlab{}.
\newblock \showarticletitle{In search of the optimal Walsh-Hadamard Transform}.
\newblock \bibinfo{journal}{\emph{IEEE}} (\bibinfo{year}{2000}).
\newblock


\bibitem[\protect\citeauthoryear{Kawaguchi and Kaelbling}{Kawaguchi and
  Kaelbling}{2019}]%
        {Kawaguchi19}
\bibfield{author}{\bibinfo{person}{K. Kawaguchi} {and} \bibinfo{person}{L.~P.
  Kaelbling}.} \bibinfo{year}{2019}\natexlab{}.
\newblock \showarticletitle{Elimination of All Bad Local Minima in Deep
  Learning}.
\newblock \bibinfo{journal}{\emph{arXiv:1901.00279}} (\bibinfo{year}{2019}).
\newblock


\bibitem[\protect\citeauthoryear{Kawaguchi, Xie, Verma, and Song}{Kawaguchi
  et~al\mbox{.}}{2018}]%
        {Kawaguchi18}
\bibfield{author}{\bibinfo{person}{K. Kawaguchi}, \bibinfo{person}{B. Xie},
  \bibinfo{person}{V. Verma}, {and} \bibinfo{person}{L. Song}.}
  \bibinfo{year}{2018}\natexlab{}.
\newblock \showarticletitle{Deep Semi-Random Features for Nonlinear Function
  Approximation}.
\newblock \bibinfo{journal}{\emph{AAAI}} (\bibinfo{year}{2018}).
\newblock


\bibitem[\protect\citeauthoryear{Klarbauer, Unterthiner, and Mayr}{Klarbauer
  et~al\mbox{.}}{2017}]%
        {Klarbauer17}
\bibfield{author}{\bibinfo{person}{G. Klarbauer}, \bibinfo{person}{T.
  Unterthiner}, {and} \bibinfo{person}{A. Mayr}.}
  \bibinfo{year}{2017}\natexlab{}.
\newblock \showarticletitle{Self-Normalizing Neural Networks}.
\newblock \bibinfo{journal}{\emph{NIPS}} (\bibinfo{year}{2017}).
\newblock


\bibitem[\protect\citeauthoryear{Lakshminarayanan, Pritzel, and
  Blundell}{Lakshminarayanan et~al\mbox{.}}{2017}]%
        {Lakshminarayanan17}
\bibfield{author}{\bibinfo{person}{B. Lakshminarayanan}, \bibinfo{person}{A.
  Pritzel}, {and} \bibinfo{person}{C. Blundell}.}
  \bibinfo{year}{2017}\natexlab{}.
\newblock \showarticletitle{Simple and Scalable Predictive Uncertainty
  Estimation using Deep Ensembles}.
\newblock \bibinfo{journal}{\emph{NIPS}} (\bibinfo{year}{2017}).
\newblock


\bibitem[\protect\citeauthoryear{Le, Sarl\'os, and Smola}{Le
  et~al\mbox{.}}{2013}]%
        {Le14}
\bibfield{author}{\bibinfo{person}{Q. Le}, \bibinfo{person}{T. Sarl\'os}, {and}
  \bibinfo{person}{A. Smola}.} \bibinfo{year}{2013}\natexlab{}.
\newblock \showarticletitle{Fastfood - Approximating Kernel Expansions in
  Loglinear Time}.
\newblock \bibinfo{journal}{\emph{ICML}} (\bibinfo{year}{2013}).
\newblock


\bibitem[\protect\citeauthoryear{Lecun, Bottou, Bengio, and Haffner}{Lecun
  et~al\mbox{.}}{1998}]%
        {Lecun98}
\bibfield{author}{\bibinfo{person}{Y. Lecun}, \bibinfo{person}{L. Bottou},
  \bibinfo{person}{Y. Bengio}, {and} \bibinfo{person}{P. Haffner}.}
  \bibinfo{year}{1998}\natexlab{}.
\newblock \showarticletitle{Gradient-based learning applied to document
  recognition}.
\newblock \bibinfo{journal}{\emph{Proceedings of the Institute of Electrical
  and Electronics Engineers}} (\bibinfo{year}{1998}).
\newblock


\bibitem[\protect\citeauthoryear{Li, Dai, Tan, Xu, and Gool}{Li
  et~al\mbox{.}}{2016}]%
        {Li16}
\bibfield{author}{\bibinfo{person}{W. Li}, \bibinfo{person}{D. Dai},
  \bibinfo{person}{M. Tan}, \bibinfo{person}{D. Xu}, {and} \bibinfo{person}{L.
  Gool}.} \bibinfo{year}{2016}\natexlab{}.
\newblock \showarticletitle{Fast Algorithms for Linear and Kernel SVM+}.
\newblock \bibinfo{journal}{\emph{CVPR}} (\bibinfo{year}{2016}).
\newblock


\bibitem[\protect\citeauthoryear{Liang, Sun, Lee, and Srikant}{Liang
  et~al\mbox{.}}{2018}]%
        {Liang18}
\bibfield{author}{\bibinfo{person}{S. Liang}, \bibinfo{person}{R. Sun},
  \bibinfo{person}{J.~D. Lee}, {and} \bibinfo{person}{R. Srikant}.}
  \bibinfo{year}{2018}\natexlab{}.
\newblock \showarticletitle{Adding One Neuron Can Eliminate All Bad Local
  Minima}.
\newblock \bibinfo{journal}{\emph{NIPS}} (\bibinfo{year}{2018}).
\newblock


\bibitem[\protect\citeauthoryear{Liao and Poggio}{Liao and Poggio}{2017}]%
        {Liao17}
\bibfield{author}{\bibinfo{person}{Q. Liao} {and} \bibinfo{person}{T. Poggio}.}
  \bibinfo{year}{2017}\natexlab{}.
\newblock \showarticletitle{Theory {II}: Landscape of the Empirical Risk in
  Deep Learning}.
\newblock \bibinfo{journal}{\emph{CBMM Memo No. 066}} (\bibinfo{year}{2017}).
\newblock


\bibitem[\protect\citeauthoryear{Lu, Dhillon, Foster, and Ungar}{Lu
  et~al\mbox{.}}{2013}]%
        {Lu13}
\bibfield{author}{\bibinfo{person}{Y. Lu}, \bibinfo{person}{P.~S. Dhillon},
  \bibinfo{person}{D. Foster}, {and} \bibinfo{person}{L. Ungar}.}
  \bibinfo{year}{2013}\natexlab{}.
\newblock \showarticletitle{Faster Ridge Regression via the Subsampled
  Randomized Hadamard Transform}.
\newblock \bibinfo{journal}{\emph{NIPS}} (\bibinfo{year}{2013}).
\newblock


\bibitem[\protect\citeauthoryear{Luong, Pham, and Manning}{Luong
  et~al\mbox{.}}{2015}]%
        {Luong15}
\bibfield{author}{\bibinfo{person}{M. Luong}, \bibinfo{person}{H. Pham}, {and}
  \bibinfo{person}{C.~D. Manning}.} \bibinfo{year}{2015}\natexlab{}.
\newblock \showarticletitle{Effective Approaches to Attention-based Neural
  Machine Translation}.
\newblock \bibinfo{journal}{\emph{EMNLP}} (\bibinfo{year}{2015}).
\newblock


\bibitem[\protect\citeauthoryear{Maas, Hannun, and Ng}{Maas
  et~al\mbox{.}}{2013}]%
        {Maas13}
\bibfield{author}{\bibinfo{person}{A.~L. Maas}, \bibinfo{person}{A.~Y. Hannun},
  {and} \bibinfo{person}{A.~Y. Ng}.} \bibinfo{year}{2013}\natexlab{}.
\newblock \showarticletitle{Rectifier Nonlinearities Improve Neural Network
  Acoustic Models}.
\newblock \bibinfo{journal}{\emph{ICML}} (\bibinfo{year}{2013}).
\newblock


\bibitem[\protect\citeauthoryear{Mhaskar, Liao, and Poggio}{Mhaskar
  et~al\mbox{.}}{2017}]%
        {Mhaskar17}
\bibfield{author}{\bibinfo{person}{H. Mhaskar}, \bibinfo{person}{Q. Liao},
  {and} \bibinfo{person}{T. Poggio}.} \bibinfo{year}{2017}\natexlab{}.
\newblock \showarticletitle{When and Why Are Deep Networks Better than Shallow
  Ones?}
\newblock \bibinfo{journal}{\emph{AAAI}} (\bibinfo{year}{2017}).
\newblock


\bibitem[\protect\citeauthoryear{Moczulski, Denil, Appleyard, and
  Freitas}{Moczulski et~al\mbox{.}}{2016}]%
        {Moczulski16}
\bibfield{author}{\bibinfo{person}{M. Moczulski}, \bibinfo{person}{M. Denil},
  \bibinfo{person}{J. Appleyard}, {and} \bibinfo{person}{N. Freitas}.}
  \bibinfo{year}{2016}\natexlab{}.
\newblock \showarticletitle{{ACDC}: A Structured Efficient Linear Layer}.
\newblock \bibinfo{journal}{\emph{ICLR}} (\bibinfo{year}{2016}).
\newblock


\bibitem[\protect\citeauthoryear{Ouyang and Cham}{Ouyang and Cham}{2010}]%
        {Ouyang10}
\bibfield{author}{\bibinfo{person}{W. Ouyang} {and} \bibinfo{person}{W.~K.
  Cham}.} \bibinfo{year}{2010}\natexlab{}.
\newblock \showarticletitle{Fast Algorithm for Walsh Hadamard Transform on
  Sliding Windows}.
\newblock \bibinfo{journal}{\emph{TPAMI}} (\bibinfo{year}{2010}).
\newblock


\bibitem[\protect\citeauthoryear{Poggio, Mhaskar, Rosasco, Miranda, and
  Liao}{Poggio et~al\mbox{.}}{2017}]%
        {Poggio17}
\bibfield{author}{\bibinfo{person}{T. Poggio}, \bibinfo{person}{H. Mhaskar},
  \bibinfo{person}{L. Rosasco}, \bibinfo{person}{B. Miranda}, {and}
  \bibinfo{person}{Q. Liao}.} \bibinfo{year}{2017}\natexlab{}.
\newblock \showarticletitle{Why and when can deep-but not shallow-networks
  avoid the curse of dimensionality: A review}.
\newblock \bibinfo{journal}{\emph{International Journal of Automation and
  Computing}} (\bibinfo{year}{2017}).
\newblock


\bibitem[\protect\citeauthoryear{Poggio and Smale}{Poggio and Smale}{2003}]%
        {Poggio03}
\bibfield{author}{\bibinfo{person}{T. Poggio} {and} \bibinfo{person}{S.
  Smale}.} \bibinfo{year}{2003}\natexlab{}.
\newblock \showarticletitle{The Mathematics of Learning: Dealing with Data}.
\newblock \bibinfo{journal}{\emph{Notices of the American Mathematical Society
  (AMS)}} (\bibinfo{year}{2003}).
\newblock


\bibitem[\protect\citeauthoryear{Rahimi and Recht}{Rahimi and Recht}{2007}]%
        {Rahimi07}
\bibfield{author}{\bibinfo{person}{A. Rahimi} {and} \bibinfo{person}{B.
  Recht}.} \bibinfo{year}{2007}\natexlab{}.
\newblock \showarticletitle{Random Features for Large-Scale Kernel Machines}.
\newblock \bibinfo{journal}{\emph{NIPS}} (\bibinfo{year}{2007}).
\newblock


\bibitem[\protect\citeauthoryear{Reddi, Ramdas, Singh, Poczos, and
  Wasserman}{Reddi et~al\mbox{.}}{2015}]%
        {Reddi15}
\bibfield{author}{\bibinfo{person}{S. Reddi}, \bibinfo{person}{A. Ramdas},
  \bibinfo{person}{A. Singh}, \bibinfo{person}{B. Poczos}, {and}
  \bibinfo{person}{L. Wasserman}.} \bibinfo{year}{2015}\natexlab{}.
\newblock \showarticletitle{On the high-dimensional power of a linear-time two
  sample test under mean-shift alternatives}.
\newblock \bibinfo{journal}{\emph{AISTATS}} (\bibinfo{year}{2015}).
\newblock


\bibitem[\protect\citeauthoryear{Rudi and Rosasco}{Rudi and Rosasco}{2017}]%
        {Rudi17}
\bibfield{author}{\bibinfo{person}{A. Rudi} {and} \bibinfo{person}{L.
  Rosasco}.} \bibinfo{year}{2017}\natexlab{}.
\newblock \showarticletitle{Generalization Properties of Learning with Random
  Features}.
\newblock \bibinfo{journal}{\emph{NIPS}} (\bibinfo{year}{2017}).
\newblock


\bibitem[\protect\citeauthoryear{Sharmanska, Quadrianto, and
  Lampert}{Sharmanska et~al\mbox{.}}{2013}]%
        {Sharmanska13}
\bibfield{author}{\bibinfo{person}{V. Sharmanska}, \bibinfo{person}{N.
  Quadrianto}, {and} \bibinfo{person}{C.~H. Lampert}.}
  \bibinfo{year}{2013}\natexlab{}.
\newblock \showarticletitle{Learning to Rank Using Privileged Information}.
\newblock \bibinfo{journal}{\emph{ICCV}} (\bibinfo{year}{2013}).
\newblock


\bibitem[\protect\citeauthoryear{Sohl-Dickstein and Kawaguchi}{Sohl-Dickstein
  and Kawaguchi}{2019}]%
        {Sohl-Dickstein19}
\bibfield{author}{\bibinfo{person}{J. Sohl-Dickstein} {and} \bibinfo{person}{K.
  Kawaguchi}.} \bibinfo{year}{2019}\natexlab{}.
\newblock \showarticletitle{Eliminating all bad Local Minima from Loss
  Landscapes without even adding an Extra Unit}.
\newblock \bibinfo{journal}{\emph{arXiv:1901.03909}} (\bibinfo{year}{2019}).
\newblock


\bibitem[\protect\citeauthoryear{Srivastava, Hinton, Krizhevsky, Sutskever, and
  Salakhutdinov}{Srivastava et~al\mbox{.}}{2014}]%
        {Srivastava14}
\bibfield{author}{\bibinfo{person}{N. Srivastava}, \bibinfo{person}{G. Hinton},
  \bibinfo{person}{A. Krizhevsky}, \bibinfo{person}{I. Sutskever}, {and}
  \bibinfo{person}{R. Salakhutdinov}.} \bibinfo{year}{2014}\natexlab{}.
\newblock \showarticletitle{Dropout: a simple way to prevent neural networks
  from overfitting}.
\newblock \bibinfo{journal}{\emph{JMLR}} (\bibinfo{year}{2014}).
\newblock


\bibitem[\protect\citeauthoryear{Tran, Pham, Carneiro, Palmer, and Reid}{Tran
  et~al\mbox{.}}{2017}]%
        {Tran17}
\bibfield{author}{\bibinfo{person}{T. Tran}, \bibinfo{person}{T. Pham},
  \bibinfo{person}{G. Carneiro}, \bibinfo{person}{L. Palmer}, {and}
  \bibinfo{person}{I. Reid}.} \bibinfo{year}{2017}\natexlab{}.
\newblock \showarticletitle{A Bayesian Data Augmentation Approach for Learning
  Deep Models}.
\newblock \bibinfo{journal}{\emph{NIPS}} (\bibinfo{year}{2017}).
\newblock


\bibitem[\protect\citeauthoryear{Vapnik and Izmailov}{Vapnik and
  Izmailov}{2015}]%
        {Vapnik15}
\bibfield{author}{\bibinfo{person}{V. Vapnik} {and} \bibinfo{person}{R.
  Izmailov}.} \bibinfo{year}{2015}\natexlab{}.
\newblock \showarticletitle{Learning Using Privileged Information: Similarity
  Control and Knowledge Transfer}.
\newblock \bibinfo{journal}{\emph{JMLR}} (\bibinfo{year}{2015}).
\newblock


\bibitem[\protect\citeauthoryear{Vapnik and Izmailov}{Vapnik and
  Izmailov}{2018}]%
        {Vapnik18}
\bibfield{author}{\bibinfo{person}{V. Vapnik} {and} \bibinfo{person}{R.
  Izmailov}.} \bibinfo{year}{2018}\natexlab{}.
\newblock \showarticletitle{Rethinking statistical learning theory: learning
  using statistical invariants}.
\newblock \bibinfo{journal}{\emph{Machine Learning}} (\bibinfo{year}{2018}).
\newblock


\bibitem[\protect\citeauthoryear{Vapnik and Vashist}{Vapnik and
  Vashist}{2009}]%
        {Vapnik09}
\bibfield{author}{\bibinfo{person}{V. Vapnik} {and} \bibinfo{person}{A.
  Vashist}.} \bibinfo{year}{2009}\natexlab{}.
\newblock \showarticletitle{A new learning paradigm: Learning using privileged
  information}.
\newblock \bibinfo{journal}{\emph{Neural Networks}} (\bibinfo{year}{2009}).
\newblock


\bibitem[\protect\citeauthoryear{Vaswani, Shazeer, Parmar, Uszkoreit, Jones,
  Gomez, Kaiser, and Polosukhin}{Vaswani et~al\mbox{.}}{2017}]%
        {Vaswani17}
\bibfield{author}{\bibinfo{person}{A. Vaswani}, \bibinfo{person}{N. Shazeer},
  \bibinfo{person}{N. Parmar}, \bibinfo{person}{J. Uszkoreit},
  \bibinfo{person}{L. Jones}, \bibinfo{person}{A.~N. Gomez},
  \bibinfo{person}{L. Kaiser}, {and} \bibinfo{person}{I. Polosukhin}.}
  \bibinfo{year}{2017}\natexlab{}.
\newblock \showarticletitle{Attention Is All You Need}.
\newblock \bibinfo{journal}{\emph{NIPS}} (\bibinfo{year}{2017}).
\newblock


\bibitem[\protect\citeauthoryear{Wang, Liu, and Huang}{Wang
  et~al\mbox{.}}{2017}]%
        {Wang17}
\bibfield{author}{\bibinfo{person}{M. Wang}, \bibinfo{person}{Y. Liu}, {and}
  \bibinfo{person}{Z. Huang}.} \bibinfo{year}{2017}\natexlab{}.
\newblock \showarticletitle{Large Margin Object Tracking With Circulant Feature
  Maps}.
\newblock \bibinfo{journal}{\emph{CVPR}} (\bibinfo{year}{2017}).
\newblock


\bibitem[\protect\citeauthoryear{Wang and Ji}{Wang and Ji}{2015}]%
        {Wang15}
\bibfield{author}{\bibinfo{person}{Z. Wang} {and} \bibinfo{person}{Q. Ji}.}
  \bibinfo{year}{2015}\natexlab{}.
\newblock \showarticletitle{Classifier Learning with Hidden Information}.
\newblock \bibinfo{journal}{\emph{CVPR}} (\bibinfo{year}{2015}).
\newblock


\bibitem[\protect\citeauthoryear{Wigner}{Wigner}{1960}]%
        {Wigner60}
\bibfield{author}{\bibinfo{person}{E. Wigner}.}
  \bibinfo{year}{1960}\natexlab{}.
\newblock \showarticletitle{The unreasonable effectiveness of mathematics in
  the natural sciences}.
\newblock \bibinfo{journal}{\emph{Communications in Pure and Applied
  Mathematics}} (\bibinfo{year}{1960}).
\newblock


\bibitem[\protect\citeauthoryear{Wilson, Hu, Salakhutdinov, and Xing}{Wilson
  et~al\mbox{.}}{2016}]%
        {Wilson16}
\bibfield{author}{\bibinfo{person}{A.~G. Wilson}, \bibinfo{person}{Z. Hu},
  \bibinfo{person}{R. Salakhutdinov}, {and} \bibinfo{person}{E.~P. Xing}.}
  \bibinfo{year}{2016}\natexlab{}.
\newblock \showarticletitle{Stochastic variational deep kernel learning}.
\newblock \bibinfo{journal}{\emph{NIPS}} (\bibinfo{year}{2016}).
\newblock


\bibitem[\protect\citeauthoryear{Xiao, Rasul, and Vollgraf}{Xiao
  et~al\mbox{.}}{2017}]%
        {Xiao17}
\bibfield{author}{\bibinfo{person}{H. Xiao}, \bibinfo{person}{K. Rasul}, {and}
  \bibinfo{person}{R. Vollgraf}.} \bibinfo{year}{2017}\natexlab{}.
\newblock \showarticletitle{FASHION-MNIST: a Novel Image Dataset for
  Benchmarking Machine Learning Algorithms}.
\newblock \bibinfo{journal}{\emph{arXiv:1708.07747}} (\bibinfo{year}{2017}).
\newblock


\bibitem[\protect\citeauthoryear{Yang, Moczulski, Denil, Freitas, Smola, Song,
  and Wang}{Yang et~al\mbox{.}}{2015}]%
        {Yang15}
\bibfield{author}{\bibinfo{person}{Z. Yang}, \bibinfo{person}{M. Moczulski},
  \bibinfo{person}{M. Denil}, \bibinfo{person}{N. Freitas}, \bibinfo{person}{A.
  Smola}, \bibinfo{person}{L. Song}, {and} \bibinfo{person}{Z. Wang}.}
  \bibinfo{year}{2015}\natexlab{}.
\newblock \showarticletitle{Deep Fried Convnets}.
\newblock \bibinfo{journal}{\emph{ICCV}} (\bibinfo{year}{2015}).
\newblock


\bibitem[\protect\citeauthoryear{Yang, Smola, Song, and Wilson}{Yang
  et~al\mbox{.}}{2014}]%
        {Yang14}
\bibfield{author}{\bibinfo{person}{Z. Yang}, \bibinfo{person}{A. Smola},
  \bibinfo{person}{L. Song}, {and} \bibinfo{person}{A.~G. Wilson}.}
  \bibinfo{year}{2014}\natexlab{}.
\newblock \showarticletitle{{\`A} la Carte - Learning Fast Kernels}.
\newblock \bibinfo{journal}{\emph{AISTATS}} (\bibinfo{year}{2014}).
\newblock


\bibitem[\protect\citeauthoryear{Zhang, Shao, and Salakhutdinov}{Zhang
  et~al\mbox{.}}{2018}]%
        {Zhang18}
\bibfield{author}{\bibinfo{person}{H. Zhang}, \bibinfo{person}{J. Shao}, {and}
  \bibinfo{person}{R. Salakhutdinov}.} \bibinfo{year}{2018}\natexlab{}.
\newblock \showarticletitle{Deep Neural Networks with Multi-Branch
  Architectures Are Less Non-Convex}.
\newblock \bibinfo{journal}{\emph{arXiv:1806.01845}} (\bibinfo{year}{2018}).
\newblock


\end{thebibliography}
